\documentclass[final,journal]{IEEEtran}

\usepackage{cite}
\usepackage[numbers]{natbib}
\usepackage[autostyle]{csquotes}

\usepackage[pdftex]{graphicx}
\graphicspath{{./figures/}}
\DeclareGraphicsExtensions{.pdf,.jpeg,.png}

\usepackage{amssymb}
\usepackage{amsmath}
\usepackage{algorithmic}
\usepackage{array}
\usepackage{multirow}

\usepackage{url}
\usepackage{hyperref}

\usepackage{xcolor, colortbl} 
\usepackage{array}
\newcommand{\mcol}{\multicolumn}

\usepackage[colorinlistoftodos]{todonotes}

\usepackage[binary-units=true]{siunitx}

\usepackage[capitalise]{cleveref}

\newcommand{\predicate}[1]{{\fontfamily{cmtt}\selectfont \small{`#1'}}}
\newcommand{\scriptpred}[1]{{\fontfamily{cmtt}\selectfont \scriptsize{`#1'}}}

\DeclareMathOperator*{\argmax}{arg\,max}

\newcommand{\andreasskip}[1]{{\color{blue}(skipped text)}}

\newcommand{\pedroskip}[1]{{\color{green}(skipped text)}}

\usepackage{textcomp}
\newcommand{\ftextnumero}{{\fontfamily{txr}\selectfont \textnumero}}

\usepackage{amsthm}
\usepackage{scrextend}
\newtheoremstyle{style}
  {\topsep} 
  {\topsep} 
  {\addtolength{\leftskip}{1em}\addtolength{\rightskip}{1em}} 
  {.0em} 
  {\bfseries} 
  {:} 
  {.3em} 
  {} 
\theoremstyle{style} \newtheorem{example}{Example}
\theoremstyle{style} 
\theoremstyle{style} 

\begin{document}


\title{Semantic Relational Object Tracking}


\author{Andreas~Persson,
        Pedro~Zuidberg~Dos~Martires,
        Amy~Loutfi,
        and~Luc~De~Raedt
\thanks{A. Persson and A. Loutfi are with the Center for Applied Autonomous Sensor Systems (AASS), Dept. of Science and Technology, {\"O}rebro University, 701 82 {\"O}rebro, Sweden.}
\thanks{P. Zuidberg Dos Martires and L. De Raedt are with Declaratieve Talen en Artificiele Intelligentie (DTAI), Department of Computer Science, KU Leuven, 3001 Heverlee, Belgium.}
\thanks{Manuscript ... July 15, 2018}}

\markboth{IEEE TRANSACTIONS ON COGNITIVE AND DEVELOPMENTAL SYSTEMS, }{Persson \MakeLowercase{\textit{et al.}}: Relational Object Tracking}


%


\IEEEoverridecommandlockouts
\IEEEpubid{\makebox[\columnwidth]{978-1-5386-5541-2/18/\$31.00~\copyright2019 IEEE \hfill} \hspace{\columnsep}\makebox[\columnwidth]{ }}
\maketitle

\begin{abstract}

This paper addresses the topic of semantic world modeling by conjoining probabilistic reasoning and object anchoring. The proposed approach uses a so-called bottom-up object anchoring method that relies on the rich continuous data from  perceptual sensor data. A novel anchoring matching function method learns to maintain object entities in space and time and is validated using a large set of trained humanly annotated ground truth data of real-world objects. For more complex scenarios, a high-level probabilistic object tracker has been integrated with the anchoring framework and handles the tracking of occluded objects via reasoning about the state of unobserved objects. We demonstrate the performance of our integrated approach through scenarios such as the shell game scenario, where we illustrate how anchored objects are retained by preserving relations through probabilistic reasoning. 
\end{abstract}


%



\section{Introduction} 
\label{section:introduction}

\begin{addmargin}[2em]{2em}
\vspace{-4mm} 
\textit{
\IEEEPARstart{C}{onsider} the classical shell game where a ball is hidden under one of three identical cups. The performer of the game rapidly moves the cups and the task of the observer is to follow the movement of the cups and to identify under which cup the ball is located. For an observer to successfully identify the right cup, he/she must successfully handle a number of subtasks. First, despite that each of the cups are visually similar, the observer must create an individual notion of each cup as a unique object so that it can be identified (e.g., "the cup in the middle"). Likewise, the observer must recognize the ball as a unique object. Secondly, even though the ball is hidden under one of the cups, the observer makes the assumption that although the ball is not perceived it should still be present under the cup. Third, as the performer rapidly moves the cups, the observer should track the cup under which the ball is hidden. And finally, the observer also needs to realize that cups can contain balls, and therefore as the cup moves, so does the ball. Depending on the level of skill of the performer (and perhaps some additional tricks) the shell game can be a difficult one to solve.
} 
\vspace{2mm} 
\end{addmargin}

Imagine, how could an autonomous agent handle this task as the observer? For this to be achieved, autonomous agents in real-world scenarios need to maintain consonance between the perceived world (through sensory capabilities) and their internal representation of the world. One way to contend with this challenge is to create a \textit{semantic world model} (i.e., a semantic object centered model of the world). As discussed by \cite{elfring.et.al-2013}, a semantic world model is a way to represent objects not only by their numeric properties such as a position vector, but also by semantic properties which share a meaning with humans. Moreover, such an internal model of objects could also be used to obtain additional facts about objects and their affordances (e.g., cups can contain balls).

In this paper, we present an approach to create a semantic world model where object entities are maintained and tracked over time. We further integrate a probabilistic reasoning component which is able to support the tracking of objects in case of occlusions due to limitation in perception or due to interactions with other objects. Our approach is based on perceptual anchoring which, traditionally, has been considered as a special case of \textit{symbol grounding}~\cite{coradeschi.et.al-2013}, used within the mobile robotics community. 
Perceptual anchoring, by definition, handles the problem to create and maintain, in time and space, the correspondence between symbols and sensor data that refer to the same physical object in the external world~\cite{coradeschi&saffiotti-2000}. 
This problem has, subsequently, been defined as the \textit{anchoring problem}~\cite{coradeschi&saffiotti-2003}.
We use \textit{bottom-up} anchoring~\cite{loutfi.et.al-2005}, whereby anchors (object representations) can be created by perceptual observations derived from interactions with the environment. For a practicable bottom-up anchoring system it is, however, essential to have a robust anchoring matching function that accurately matches perceptual observations against the perceptual data of previously maintained anchors. For this purpose, we introduce a novel method that replaces a traditionally hand-coded anchoring matching function by a learned model (c.f. \cite{persson.et.al-2017}).

Furthermore, we integrate our bottom-up anchoring framework with a probabilistic reasoning system, Dynamic Distributional Clauses (DDC)\cite{Nitti:2016:PLP:2949339.2949375}, to facilitate reasoning about object entities at a symbolic level. DDC is an extension of the probabilistic logic programming language ProbLog~\cite{fierens2015inference}, which can handle continuous random variables in addition to discrete ones. This predestines DDC to be utilized for reasoning within robotics, where the world is inherently continuous and uncertain.
Integrating a reasoning system into the anchoring framework allows us to dynamically feed back information to the anchoring system and update the retained semantic world model with information stemming from our probabilistic model of the world. The novelty of our approach is that we encode the affordance of objects at a high level of abstraction. This allows us to not only track objects but to also represent the interaction between objects. A standard object tracker could handle occlusions, but in cases where an object is first occluded but then moved as it has been encapsulated by another object, a standard tracker would fail. Particularly, if this type of object interaction would occur over an extended period of time. Therefore, we believe that by eventually learning affordances, we could further automate this reasoning process of objects. For now, we do not learn the affordance but, instead, encode relations in the probabilistic reasoner.

The remainder of the paper is structured as follows. In Section~\ref{section:related_work} we introduce previously presented works relevant to object anchoring in relation to probabilistic tracking, data association and reasoning. In Section~\ref{section:background}, we state the general background of used approaches and give an overview of both object anchoring and Dynamic Distributional Clauses, a probabilistic logic programming language. Our novel framework for improving object anchoring and tracking through the integration of probabilistic reasoning into object anchoring is, subsequently, presented in Section~\ref{section:method}. In Section~\ref{section:results}, we present our result of learning the anchoring matching function, utilizing human annotated data and machine learning techniques. Moreover, we introduce the proof-of-concept of how the use of probabilistic reasoning and object tracking is employed in order to preserve a consistent world model. Finally, we conclude the paper with a summary and a discussion regarding possible directions of future work, presented in Section~\ref{section:conclusions}. 


The work presented in this paper has been carried out as part of a larger project titled ReGROUND\footnote{http://reground.cs.kuleuven.be/}~\cite{reground.glu-2017}, which strives towards the greater ambition of using relational symbol grounding for the purpose of affordance learning in robotics. More specifically, the ReGROUND project hypothesizes that \emph{the grounding process should consider the full context of the environment}, which consists of multiple objects as well as their relationships and properties, and how the state of these objects changes through actions and over time. The framework that we introduce in this paper (presented in Section~\ref{section:method}), constitutes one of the corner pillars of this project -- namely, the groundwork that provides the basic data about the objects and their properties and relations. 
\section{Related Work}
\label{section:related_work}
The importance of data association and object tracking in relation to perceptual anchoring was widely discussed by \citeauthor{leblanc-2010} in his thesis on the topic of cooperative anchoring~\cite{leblanc-2010}. Around the same time, and as an alternative to traditional anchoring, early work on perceptual and probabilistic anchoring was presented by \citeauthor{blodow.et.al-2010} in \cite{blodow.et.al-2010}. The history of objects was maintained as computationally complex scene instances and the approach was, therefore, mainly intended for solving the problem of anchoring and maintaining coherent instances of objects in object kidnapping scenarios, i.e., when an object disappears from the scene and later reappears in a different location.

The idea of probabilistic anchoring was subsequently further explored by \citeauthor{elfring.et.al-2013}, which introduced probabilistic multiple hypothesis anchoring~\cite{elfring.et.al-2013}. This approach utilizes Multiple Hypothesis Tracking-based data association (MHT)\cite{reid-1979}, in order to maintain changes in anchored objects, and thus, maintain an adaptable world model. In similarity with their work, we acknowledge that a proper data association is important for object anchoring, and we support the requirements identified by the authors for a changing and adaptable world modeling algorithm, which are formulated to include: 1) \textit{appropriate anchoring}, 2) \textit{data association}, 3) \textit{model-based object tracking}, and 4) \textit{real-time execution}. Contrary to the work of \citeauthor{elfring.et.al-2013}, however, we address the tasks of appropriate anchoring and data association in a holistic fashion by introducing a learned anchoring matching function that administers both tasks. Moreover, in contrast to the work presented in \cite{elfring.et.al-2013}, we do not encourage a highly integrated approach that supports a tight coupling between object anchoring/probabilistic data association and object tracking. Instead, our approach maintains a loose coupling, which is motivated by the fact that a MHT procedure will inevitably suffer from the \textit{curse of dimensionality}~\cite{bellman-1957}. A purely \textit{probabilistic anchoring} approach, as presented in \cite{elfring.et.al-2013}, will, therefore, further propagate the curse of dimensionality into the concept of anchoring.

The limitation in the use of MHT for world modeling has also been acknowledged in a recent publication on the topic of data association for semantic world modeling~\cite{wong.et.al-2015}. While this work inherits the same problem formulation, it substantially differs in approach. The authors discuss and exemplify issues related to the use of a tracking-based approach for world modeling, such as intractable branching of the tree-structured tracks of possible hypothesis, and instead, suggests a \textit{clustering} approach based on Markov Chain Monte Carlo Data Association (MCMCDA)\cite{oh.et.al-2009}. In the same work on data association for semantic world modeling, \citeauthor{wong.et.al-2015} also pointed out some characteristics that differentiate world modeling from target tracking. For example, an appropriate world modeling algorithm should take into consideration that the state of most objects do not change between frames, while a tracking algorithm must consider unchanged objects as possible valid targets. For the approach presented in this work, we are assuming similar characteristics through high-level tracking of objects that are not directly perceived by the sensory input data.

From the relational point of view, which enables us to carry out reasoning, some research has been conducted on utilizing relations to improve the tracking of real world entities and state estimation (\citep{manfredotti2009modeling, mosenlechner2009using, tenorth2013knowrob, nitti2014relational}). The most expressive of theses approaches is by \citeauthor{nitti2014relational} in \citep{nitti2014relational}. They utilize a relational particle filter, expressed in DDC, to carry out the tracking of objects and handle occlusions. In a box packaging scenario, where boxes are placed inside each other, they showed that binary predicates like $\mathtt{inside}/2$ are helpful when tracking objects that are not directly observable. However, ~\citeauthor{nitti2014relational} assumed the data association problem to be solved by identifying the objects (boxes) by augmented reality (AR) tags -- hence, very strictly limiting the usage of their framework in real-world scenarios.


\section{Background}
\label{section:background}

In this section, we outline the general background of our used methods.
We will both present the traditional definition found within the concept of \textit{perceptual anchoring}, as well as an overview of \textit{Dynamic Distributional Clauses}. We will further outline the problem domain of existing definitions and methods that address the problem of object anchoring. The same outline provides the motivation for our suggested approach, presented in subsequent Section~\ref{section:method}.

\subsection{Background on Perceptual Anchoring}
\label{section:background_anchoring}

Perceptual anchoring, originally presented by \citeauthor{coradeschi&saffiotti-2000} in \cite{coradeschi&saffiotti-2000}, has been defined in an attempt to address a subset of the symbol grounding applied to robotics and intelligent systems. The notion of perceptual anchoring has undergone several extensions and refinements since its first definition. Some notable works in this regard include the addition of bottom-up anchoring \cite{loutfi.et.al-2005}, extensions for multi-agent systems \cite{leblanc&saffiotti-2008}, considerations for non-traditional sensing modalities and knowledge based anchoring given full scale knowledge representation and reasoning systems \cite{loutfi-2006, loutfi&coradeschi-2006, loutfi.et.al-2008}, and integration of large-scale knowledge bases together with of common-sense reasoning in distributed anchoring \cite{daoutis.et.al-2012}. All those works have been presented with a common ingredient of a number of prerequisite components from~\cite{coradeschi&saffiotti-2000}, including: a \emph{symbolic system} (including: a set $\mathcal{X} = \{ x_1, x_2,\dots \}$ of individual symbols (variables and constants); a set $\mathcal{P} = \{ p_1, p_2,\dots \}$ of predicate symbols), a \emph{perceptual system} (including: a set $\Pi = \{ \pi_1, \pi_2,\dots \}$ of percepts; a set $ \Phi = \{ \phi_1, \phi_2,\dots \}$ of attributes), and \emph{predicate grounding relation} $g \subseteq \mathcal{P} \times \Phi \times D(\Phi)$, which encodes the correspondence between (unary) predicates and values of measurable attributes. The relation $g$ maps a certain predicate to compatible attribute values. An overview of the anchoring components and their relations is exemplified in Example~\ref{ex:anchoring_components}.

\begin{figure}[!ht]
	\begin{center}
		\includegraphics[width=0.46\textwidth]{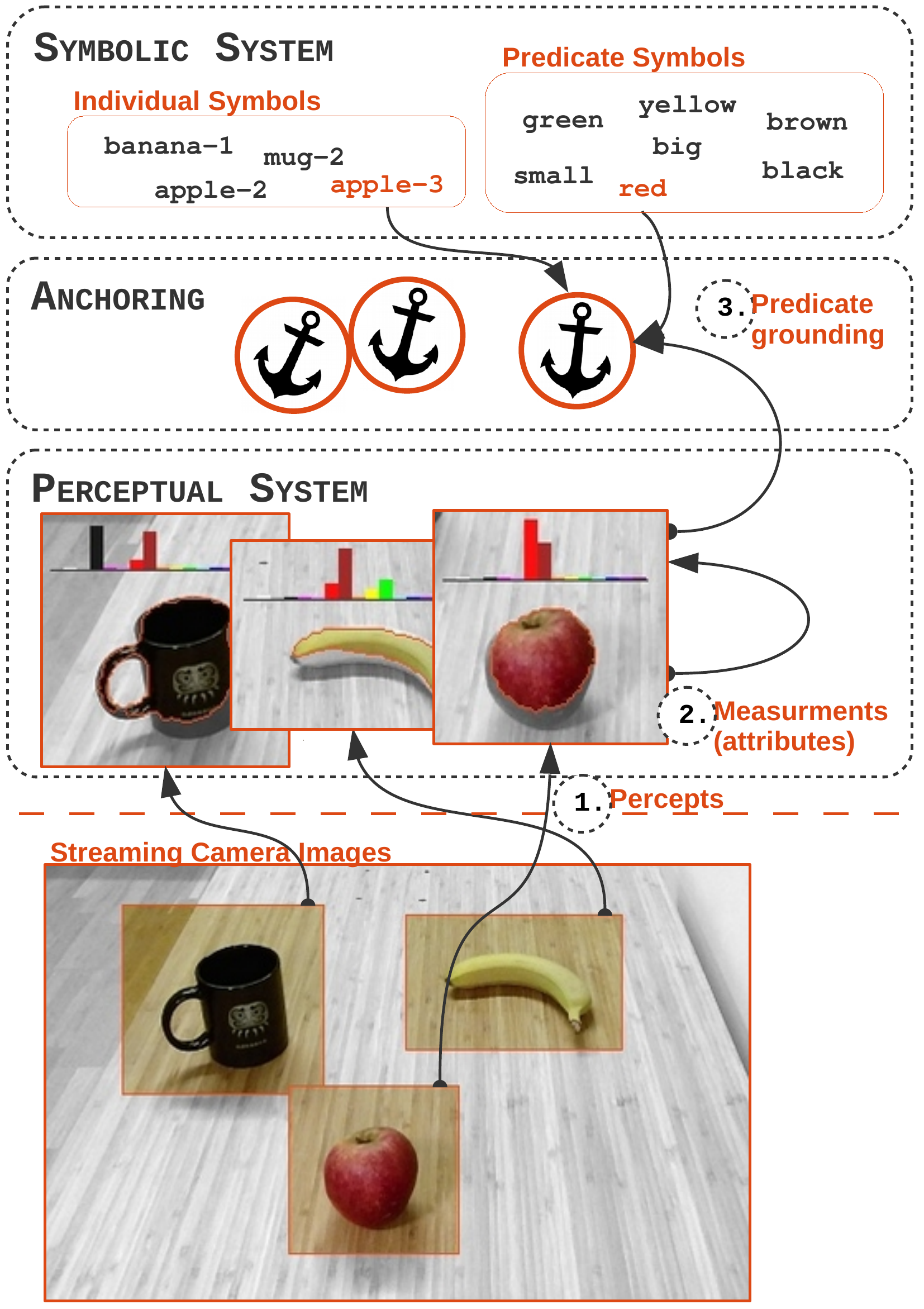}
	\end{center}
	\caption{A graphical illustration of the anchoring components and their interconnections. }\label{fig:anchoring_components}
\end{figure}

\begin{example}\label{ex:anchoring_components}
Consider the captured camera image with segmented image regions, seen in Figure~\ref{fig:anchoring_components}. Each segmented region corresponds to an individual \textit{percept} captured by the \textit{perceptual system}, see Figure~\ref{fig:anchoring_components} -- \ftextnumero~1. We denote the percepts $\pi_1$, $\pi_2$ and $\pi_3$, which corresponds to observed physical objects \textit{banana}, \textit{apple} and \textit{mug}, respectively. For each percept is, subsequently, a number of \textit{attributes} measured, e.g., \textit{color}, \textit{size}, etc. One such attribute is a \textit{color attribute} measured as a normalized color histogram over the masked area of \textit{percept} $\pi_2$, illustrated in Figure~\ref{fig:anchoring_components} -- \ftextnumero~2. For clarity, we denote the measured \textit{color attribute} as attribute $\phi_2^{color}$, which have values in a \textit{domain} that is equal to the $n$ number of histogram bins. Finally, the \emph{predicate grounding relation} $g$, illustrated in Figure~\ref{fig:anchoring_components} -- \ftextnumero~3, for the aforementioned \textit{color attribute} can be seen as the encoded correspondence between specific peeks in the color histogram and certain \textit{predicate symbols}, e.g.:  $$g( \textnormal{\predicate{red}}, color, \argmax_{i=1 \dots n}( \phi^{color}_{2,i} ) ) ~\textbf{\textit{iff}}~ i = 6$$
\end{example} 

An anchor is, consequently, an internal data structure $\alpha^x_t$, indexed by time $t$ and identified by a unique individual symbol $x$ (e.g., \predicate{coffee-mug-1}, \predicate{box-4}, etc.), which encapsulates and maintains the correspondences between percepts and symbols that refer to the same physical object. Following the definition presented by \cite{loutfi.et.al-2005}, the principle functionalities to create and maintain anchors in a bottom-up fashion, i.e., functionalities triggered by a perceptual event, are:

\begin{itemize}
    \setlength\itemsep{0.3em}
    
	\item \textit{Acquire} -- initiates a new anchor whenever a candidate object is received that does not match any existing anchor $\alpha^x$. This functionality defines a structure $\alpha^x_t$, index by time $t$ and identified by a unique identifier $x$, which encapsulates and stores all perceptual and symbolic data of the candidate object. 

	\item \textit{Re-acquire} -- extends the definition of a matching anchor $\alpha^x$ from time $t-k$ to time $t$. This functionality assures that the percepts pointed to by the anchor are the most recent and adequate perceptual representation of the object.
 
\end{itemize}

Given the defined functionalities above, it is evident that an initial matching function is essential to determine if a candidate object is matching an existing anchor or not. In previously reported work on perceptual anchoring, the problem of matching anchors has mostly been addressed through a simplified approach based on the use of symbolic values (or left out entirely), and where the predicate grounding relation mapping between symbolic predicate values and measured attribute values are commonly facilitated by the use of conceptual spaces \cite{chella.et.al-2004}. Conceptual spaces can be thought of as a discretization of the continuous perceptual attribute space, which consequently also result in a loss of information. The procedure of creating and maintaining anchors, based on the discretized symbolic values, must accordingly either be handled by a probabilistic system, as in the case of \citep{elfring.et.al-2013}, or through the use of additional knowledge and the use of a reasoning system, as in the case of \cite{daoutis.et.al-2012}. 
In this paper, we move the matching function down to the perceptual level by presenting a novel matching approach that facilitates the rich information found within the measured attribute values, further presented in Section~\ref{section:anchoring_system}.

Moving the anchoring matching function to the perceptual level will, inevitably, also introduce another level of complexity since measured attributes must be compared based on continuous attribute values. The system must, subsequently, both recognize previously observed objects and detect (or anchor) new previously unknown objects based on the result of the initial matching function. In the case of open world scenarios without a fixed number of possible objects that the system might encounter, this is undoubtedly a challenging issue. In this paper, we address this issue and present an evaluation that addresses the problem of learning how to determine if an object has previously been perceived (or not), in the context of bottom-up perceptual anchoring, presented in Section~\ref{section:learning_anchoring}.


Traditionally (\cite{coradeschi&saffiotti-2000,coradeschi&saffiotti-2003}), there has also been a third anchoring functionality; a \textit{track} functionality\footnote{The \textit{track} functionality was formally integrated with the \textit{re-acquire} functionality for the extension to sensor-driven bottom-up anchoring \cite{loutfi.et.al-2005}, such that no distinction was made between extending the definition for an anchor from time $t-1$ or $t-k$.}. This track functionality has recently been revised and explored in the interest of integrating object tracking into the concept of anchoring, introduced in \cite{persson.et.al-2017}, which suggested a tracking functionality highly integrated with a point cloud based particle filter object tracking approach on the initial perceptual sensor level~\citep{rusu&cousins.2011}. However, the performance of the previously suggested framework was, consequently, affected by the computational load of the used object tracking approach, which requires tracking of computational demanding $3{\text -}D$ point cloud data. In this paper, we further explore the integration between object tracking and perceptual anchoring, presented in Section~\ref{section:integration}. This integration is based upon the belief that the integration should be loosely coupled in order to sustain the benefits of both the ability to maintaining individual instances on objects in a larger scale, as in the case of perceptual anchoring, as well as efficiently and logically track object instances over time, as in the case of probabilistic reasoning. In particular, we utilize \textit{Dynamic Distributed Clauses} in order to track objects on a higher conceptual level and, hence, also enable reasoning about the objects perceived on the anchoring level, as presented in Section~\ref{section:tracking}. 


\subsection{Overview of Dynamic Distributional Clauses}
\label{section:overview_ddc}

Dynamic Distributional Clauses is an extension of the logic programming language Prolog \citep{sterling1994art} that is capable of handling discrete and continuous random variables at the same time. Programs written in DDC allow for high-level (discrete variables) reasoning with low-level sensor input (continuous variables).

In logic programming, reasoning happens through the usage of symbols. These are either terms or predicates. The latter are often referred to as relations. Terms can be constants, logical variables or $n$-ary functors applied to an $n$-tuple of terms. Constants can only have one assigned value to them, which means that only one interpretation is possible. This is in contrast to logical variables, which have multiple interpretations. More concretely, a logical variable $\mathtt{X}$ is a variable ranging over the set of all possible ground terms. Lastly, we also have terms of the form $\mathtt{p}(\mathtt{t}_1,\dots,\mathtt{t}_n)$ with $\mathtt{p}/ n$ being an $n$-ary predicate and all $\mathtt{t}_i$'s being terms themselves. This last kind of terms are dubbed atoms.

In the static case, i.e., when there is no explicit dependency on time in the terms, DDC programs reduce to \textit{Distributional Clauses} (DC) \citep{gutmann2011magic, nitti2013particle} programs. A distributional clause is of the form $\mathtt{h}\sim \mathcal{D}\leftarrow \mathtt{b}_1,\dots,\mathtt{b}_n$, where $\sim$ is a predicate in infix notation and $\mathtt{b}_i$'s are literals, i.e., atoms or their negation. $\mathtt{h}$ is the name of a random variable and $\mathcal{D}$ tells us how the random variable is distributed -- both are formally terms. The meaning of such a clause is that each grounded instance of a clause $(\mathtt{h}\sim \mathcal{D}\leftarrow \mathtt{b}_1,\dots,\mathtt{b}_n)\theta$ defines a random variable $\mathtt{h}\theta$ that is distributed according to $\mathcal{D}\theta$. A grounding substitution $\theta=\{ \mathtt{V}_1 / \mathtt{t}_1, \dots, \mathtt{V}_n / \mathtt{t}_n \}$ is a transformation that simultaneously substitutes all logic variables $\mathtt{V}_i$ in a distributional clause with non-variable terms $\mathtt{t}_i$. Here we see that random variables and distributions are themselves not necessarily grounded by definition. The mean of a normal distribution can for instance depend on random variables. For the atom $\mathtt{h}\theta$ to be defined it is necessary that all atoms $\mathtt{b}_i \theta$ in the distributional clause evaluate to true. Labeling a distributional clause with time indexes allows for declaring dynamic models via defining a transition model from time step $\mathtt{t}$ to time step $\mathtt{t+1}$.

Inference in DC is conducted through importance sampling combined with backward reasoning, likelihood weighting and Rao-Blackwellization \citep{Nitti:2016:PLP:2949339.2949375}, while filtering in the dynamic case is carried out through particle filtering \citep{nitti2013particle}.

To sum up, DDC is a template language that defines conditional probabilities for discrete and continuous random variables:
$p(\mathtt{h}\theta | \mathtt{b}_1 \theta, \dots, \mathtt{b}_n \theta)=\mathcal{D}\theta$. 

\section{Novel Method: Combined Object Anchoring and Tracking Framework}
\label{section:method}

\begin{figure}[ht!]
	\begin{center}
		\includegraphics[width=0.46\textwidth]{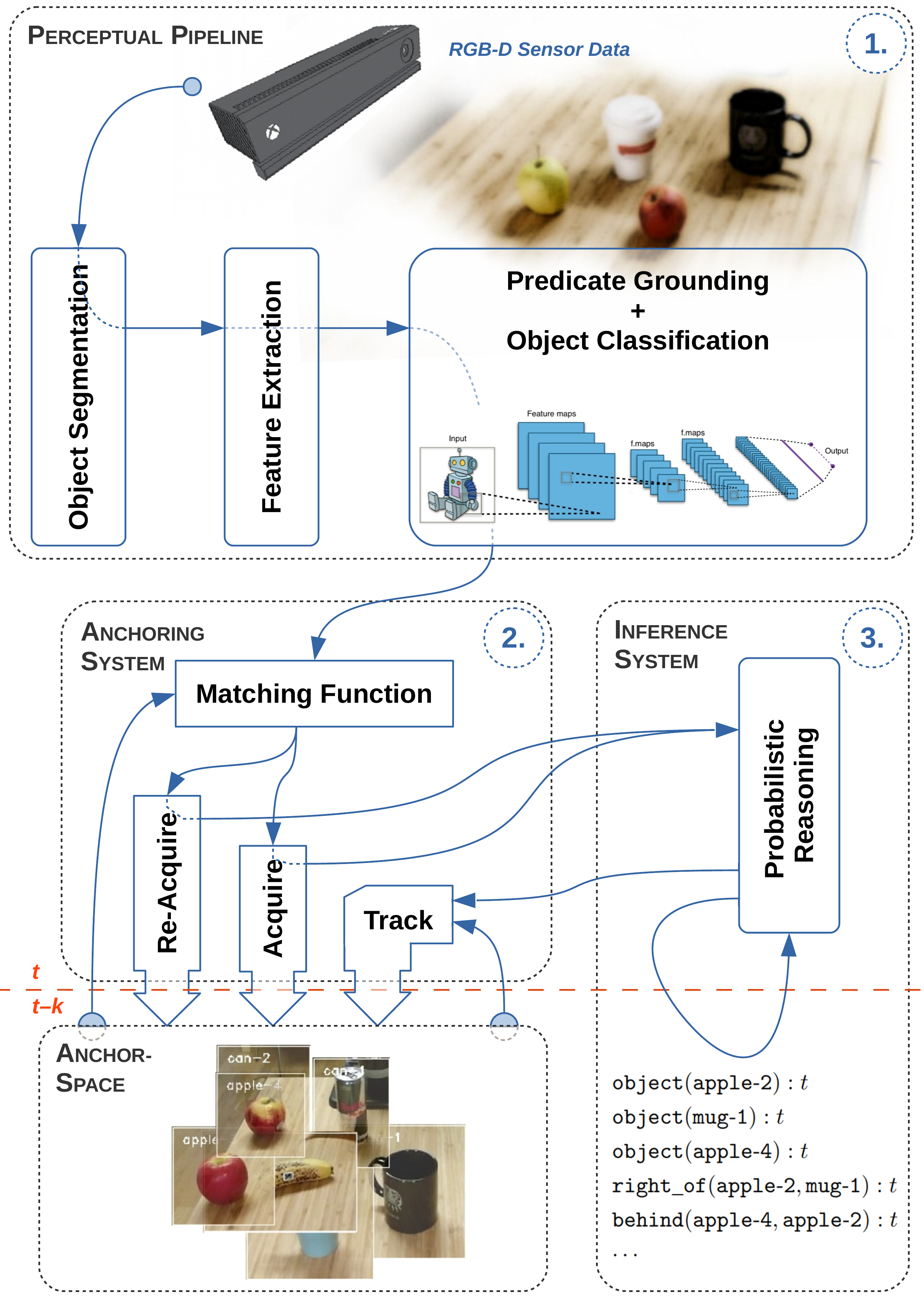}
	\end{center}
	\caption{Overview of our combined framework architecture. The overall framework is modularized architecture that consist of three core sub-systems (each described in further details in this section): \textit{1)} an initial \textit{perceptual pre-processing pipeline} with the purpose of detecting, segmenting and processing perceived objects in the physical world, \textit{2)} an \textit{anchoring system} with the purpose of creating and maintaining updated and consistent representations (anchors) of perceived objects, and \textit{3)} an \textit{inference system} with the purpose aiding the anchor system and track objects in complex dynamic scenarios. }\label{fig:system_overview}
\end{figure}

Our suggested combined framework architecture, seen in Figure~\ref{fig:system_overview}, is a modularized architecture that utilizes libraries and communication protocols available in the Robot Operating System (ROS)\footnote{http://www.ros.org/}. Each module/sub-system (described in details below) has a dedicated task, while the overall goal of the combined framework is to create and maintain coherent and accurate representations (anchors) of perceived real-world objects. The same anchor representations also provides the historical background information, i.e., information about objects last perceived at time $t-k_x$, which is used by the anchoring system to process and anchor objects perceived at the present time $t$, illustrated in Figure~\ref{fig:system_overview} -- \ftextnumero~2. Furthermore, the framework utilizes an inference system, illustrated in Figure~\ref{fig:system_overview} -- \ftextnumero~3, which aids the anchor system in complex dynamic scenes. Finally, the architecture is a sensor-driven architecture that is triggered by perceptual data, i.e. sensor readings, which initially are pre-processed by a perceptual pipeline, illustrated in Figure~\ref{fig:system_overview} --  \ftextnumero~1.

\subsection{Implementation Details: Pre-processing Pipeline}
\label{section:implementation_etails}

The overall background \textit{pre-processing pipeline}, with the goal of detecting and segmenting objects, extract features from detected objects, classifying and symbolically grounding each object instances, is illustrated in Figure~\ref{fig:system_overview} --  \ftextnumero~1. For this purpose, our system setup relies upon publicly available core libraries and systems, including: the Point Cloud Library\footnote{http://pointclouds.org/} (PCL), the Open Computer Vision library\footnote{http://opencv.org/} (OpenCV), and the Robot Operating System (ROS). It should also be noted, all methods and techniques covered in this section are considered to be replaceable \textit{black-box} approaches that are used for the means of providing background for the subsequent theoretical Section~\ref{section:theoretical_aspects}. For example, the used object segmentation method could be replaced with a convolutional network-based semantic segmentation approach~\cite{long.et.al-2015}. This requires, however, an adequate dataset of pixel-wise mask annotations for training the network to detect the objects of interest, which is something that is not always publically available and must, therefore, further be addressed, e.g. through weakly supervised learning~\cite{khoreva.e.tal-2017}. Nevertheless, the details on used techniques, for the presented architecture, are here covered for completeness and reproducibility.

More specifically, the initial step of our pre-processing pipeline is an \textit{object segmentation} method, which is performed with the purpose of detecting arbitrary objects of interest in the scene. The deployed object segmentation method is based on organized point cloud data (i.e., the organization of point cloud data is identical to the rows and columns of the imagery data from which the point cloud originates), which are given as input data by a Kinect2 RGB-D sensor. To establish the initial connection between the ROS environment and the Kinect2 sensor has further the ROS-Kinect2 bridge~\cite{iai_kinect2} been integrated and used together with the presented framework architecture. Hence, the segmentation procedure can briefly be described using the following steps: 

\begin{itemize}
  \setlength\itemsep{0.3em}
  
  \item Estimate $3{\text -}D$ surface normals based on integral images \cite{holzer.et.al-2012}. This function uses the algorithm for calculating average $3{\text -}D$ gradients over six integral images, where the horizontal and vertical $3{\text -}D$ gradients are used to compute the normal as the cross-product between two gradients.

  \item Planar segmentation based on the calculated surface normals.

  \item Object segmentation through clustering of the remaining points (points that are not part of the detected planar surfaces). This segmentation uses a connected component segmentation, presented by \cite{trevor.et.al.2013}, where a Euclidean comparison function is used to connect the components that constitute the cloud cluster of an individual object.

\end{itemize}

Moreover, provided that object instances have been segmented based on the full spectrum of available RGB-D data (as described above), we are further able to exploit the advancements in deep learning for \textit{image classification} in the final \textit{object classification} procedure of our pre-processing pipeline. The Convolutional Neural Networks (CNN) architecture used in this case is based on the $\SI{1}{K}$ GoogLeNet model, developed by \cite{szegedy.et.al-2015}, and which originally was trained on the ILSVRC 2012 visual challenge dataset~\cite{ilsvrc-2015}. For this work, we have, however, extracted a subset of the ImageNet database~\cite{deng.et.al-2009} and fine-tuned the model for $101$ objects categories that are relevant for a household domain, e.g., \predicate{mug}, \predicate{spoon}, \predicate{banana}, \predicate{tomato}, etc., where the model was trained for a \textit{top-1} accuracy of $73.4\%$ (and a \textit{top-5} accuracy of $92.0\%$).

\subsection{Theoretical Aspects: Precepts, Attributes and Symbols}
\label{section:theoretical_aspects}

The resulting output of the object segmentation is $m$ point cloud clusters (where $m$ varies between frames). For consistency with the definition of anchoring, we denote segmented clusters as percepts $\{\pi^{cloud}_1, \pi^{cloud}_2, \dots \pi^{cloud}_m\}$, which each corresponds to the spatial $3{\text -}D$ point cloud data of an individual object. Posterior to the segmentation of the point cloud clusters, the same RGB-D data is also used for segmenting corresponding visual $2{\text -}D$ imagery data of each detected object. This image segmentation is entirely based on the prior point cloud clusters and a projection between the $3{\text -}D$ point cloud frame and the $2{\text -}D$ visual RGB frame of the RGB-D sensor. Also, we denote visual data as percepts $\{\pi^{image}_1, \pi^{image}_2, \dots \pi^{image}_m\ \}$, which each corresponds to the visual $2{\text -}D$ imagery data of a segmented object.

Next, both segmented perceptual $3{\text -}D$ point cloud data and $2{\text -}D$ visual data are further forwarded to a \textit{feature extraction} procedure. The first step of this feature extraction procedure is to extract both a \textit{position attribute} as the point at the geometrical center of each segmented percept $\pi^{cloud}_y$, and a \textit{size attribute} as the $3{\text -}D$ bounding box around each percept $\pi^{cloud}_y$
, where $\pi^{cloud}_{y \mid y=1,2 \dots m} \in \{\pi^{cloud}_1, \pi^{cloud}_2, \dots \pi^{cloud}_m\ \}$. The extracted \textit{position attribute} is here denoted $\phi^{pos}_y \in \mathbb{R}^3$, while the corresponding \textit{size attribute} is denoted $\phi^{size}_y \in \mathbb{R}^3$. Furthermore, a \textit{color attribute} $\phi^{color}_y$ is extracted for each visual percept $\pi^{image}_y$, which is measured as a color histogram (in the HSV color space).

Finally, extracted attributes together with the perceptual data are further forwarded to a combined \textit{predicate grounding} and \textit{object classification} procedure. This procedure has the purpose of both grounding and associating a symbolic value with each extracted attribute, as well as classifying each object and further associating each object with an object category label. The predicate grounder is here responsible for grounding each measured attribute $\phi_y$ (of the set $\Phi_y$, that originates from the same physical object) to a predicate grounding symbol $p_y$. For example, a certain peek in a color histogram, measured as a $\phi^{color}_y$ attribute, is grounded to the symbol \predicate{red}, such that $p^{color}_y \text{=} \predicate{red}$ (cf. Example~\ref{ex:anchoring_components}). In the context of anchoring, we further assume that all trained object categories of used GoogLeNet model are part of the set of possible predicate symbols $\mathcal{P}$. The inputs for the object classification procedure are, subsequently, the segmented visual percepts $\pi^{image}_y$, while resulting object categories together with predicted category probabilities are denoted by $p^{class}_y \in \mathcal{P}$ and $\phi^{class}_y$, respectively.

\subsection{Anchoring Management}
\label{section:anchoring_system}

The entry point for the \textit{anchoring system}, seen in Figure~\ref{fig:system_overview} -- \ftextnumero~2, is a \emph{matching function}. This function assumes a bottom-up approach to perceptual anchoring, described in \cite{loutfi.et.al-2005}, where the system constantly receives candidate anchors and invokes a number of different matching algorithms (one matching algorithm for each measured attribute in the set $\Phi_y = \{\phi^{class}_y, \phi^{color}_y, \phi^{size,}_y \phi^{pos}_y \}$) in order to determine if an anchor, $\alpha_x$, has previous been perceived or not. 

\subsubsection{Matching Function}
\label{section:matching_function}

More specifically, an unknown set of attributes $\Phi_y$ is compared against the set of attributes $\Phi_x$ of an existing anchor $\alpha^x$. The combined result of all individual invoked matching algorithm determines, subsequently, if an anchored object has previously been perceived or not.
In details, a classification attribute $\phi^{class}_y$ and symbol $p^{class}_y$ of a candidate object is firstly compared against the classification attribute and symbol of a previously stored anchor according to:

\begin{align}
  & d^{class}_{x,y}( \phi^{class}_x, \phi^{class}_y ) = \nonumber \\
  & \left\{    \begin{array}{l l}
    \ \exp {\left(- \frac{ | \phi^{class}_x - \phi^{class}_y | }{ \phi^{class}_x + \phi^{class}_y } \right)}   &  \textrm{\textbf{if} $p^{class}_x \equiv p^{class}_y$} \\
    \ 0  &  \textrm{\textbf{else}}
  \end{array} 
  \right.
  \label{eq:caffe_diff}   
\end{align}

We interpret the $d^{class}$ as the exponentially decaying relative $L^1$-distance between the two attribute values $\phi_x^{class}$ and $\phi_y^{class}$. This means that we exponentially penalize the distance between two objects in the class attribute space.  

Secondly, the color histogram of a color attribute $\phi^{color}_y$ of a candidate object is compared (assuming normalized color histograms) according to the \emph{color correlation}:

%
%

\begin{align}
  & d^{color}_{x,y}( \phi^{color}_x, \phi^{color}_y ) = \nonumber \\
  &  \frac{1}{2} \left( 1 +  \frac{ \displaystyle\sum_{i=1}^{n} (\phi^{color}_{x,i} - \mu_x)(\phi^{color}_{y,i} - \mu_y) }{ \displaystyle
  \sqrt{\sum_{i=1}^{n}(\phi^{color}_{x,i} - \mu_x)^2 \displaystyle\sum_{i=1}^{n}(\phi^{color}_{y,i} - \mu_y)^2 } }  \right)
  \label{eq:color_corr} 
\end{align}
Where $n$ is the number of histogram bins and the index $i$ gives the $i$-th histogram bin value of $\phi_x^{color}$ and $\phi_y^{color}$, respectively. $\mu_x$ and $\mu_y$ are the \textit{color mean value} of each histogram, given according to:

\begin{equation*}
   \mu_x = \frac{1}{n} \sum_{i=1}^{n}\phi^{color}_{x,i}, \mu_y = \frac{1}{n} \sum_{i=1}^{n}\phi^{color}_{y,i}
   \label{eq:color_mean}   
\end{equation*}

Next, the position attribute $\phi^{pos}_y$, and the (last updated) position $\phi^{pos}_x$ of a previously stored anchor $\alpha^x$, is calculated according to the $L^2$-distance (in $3{\text -}D$ spatial space). Inspired by the work presented by \cite{blodow.et.al-2010}, this distance is then mapped to a \textit{normalized similarity distance} according to:

\begin{equation}
  d^{pos}_{x,y}( \phi^{pos}_y, \phi^{pos}_x ) =  e^{- {L^2}(\phi^{pos}_y, \phi^{pos}_x ) } 
  \label{eq:position_diff}   
\end{equation}

Furthermore, the size attribute $\phi^{size}_y$ of a candidate object is compared according to the \emph{generalized Jaccard similarity} (for the bounding boxes in $3{\text -}D$ space):

\begin{equation}
   d^{size}_{x,y}( \phi^{size}_x, \phi^{size}_y ) = \frac{
      \sum_{i=1}^{3} min(\phi^{size}_{x,i}, \phi^{size}_{y,i} ) }{
      \sum_{i=1}^{3} max(\phi^{size}_{x,i}, \phi^{size}_{y,i} ) }
   \label{eq:jaccard_index}
\end{equation}

Motivated by the importance of the time $t$ within the concept of anchoring, the difference in time since last recorded observation of a previously stored anchor $\alpha^x$, defined at $t-k_x$, is finally mapped to a similar normalized distance according to: 

\begin{equation}
  d^{time}_{x,y}( t, t-k_x ) = \frac{2}{ 1 + e^{t-(t-k_x)} } = \frac{2}{ 1 + e^{k_x} } 
  \label{eq:time_diff}   
\end{equation}

Consequently, all given matching distance values, \cref{eq:caffe_diff,eq:jaccard_index,eq:color_corr,eq:position_diff,eq:time_diff}, are given in the interval $[0.0, 1.0]$, and all distance values are, therefore, also commensurable.




\subsubsection{Creating and Maintaining Anchors} 

Combining all matching distance values, given by \cref{eq:caffe_diff,eq:jaccard_index,eq:color_corr,eq:position_diff,eq:time_diff}, and determining whether a candidate anchor has previously been perceived or not is a non-trivial task. Especially in the context of \textit{bottom-up} anchoring with both continuous possibilities of objects, and continuous similarity distances given by the initial \textit{matching function}. The matching distance values can be combined in many different ways, e.g.: through a \textit{min} or \textit{max} function, by the \textit{weighted average} with different weights, etc. Nonetheless, a threshold value is ultimately required in order to determine if the combined result is to be considered as a match (or not). In this paper, we will shed some light upon this issue and present our work on the topic of learning the anchoring matching function, which determines if an object has previously been observed or not.

At this point we would like to stress that the architecture of the anchoring system is completely agnostic towards how the matching function was learned. This means that the anchoring system considers the matching function to be an exchangeable \textit{black-box} approximation of the true anchor-percept matching. In Section~\ref{section:learning_anchoring} we compare different classifiers to each other that could be potentially used to approximate the matching.

Regardless of the used classification algorithm, the process of the anchoring is to ultimately create or maintain anchors through either one of the two principal functionalities: \textit{acquire} and \textit{re-acquire}, respectively (further described in Section~\ref{section:background_anchoring}). The \textit{anchor-space}, in which the anchors are maintained and stored, is in this case expressed as a \textit{permanent world model} (PWM). We further enhance the traditional \textit{acquire} functionality by utilizing the deep learning classifier such that a unique identifier $x$ is further generated based on the classification symbol $p^{class}$, e.g., for an object classified as a \predicate{cup}, a corresponding unique identifier could be generated as $x = \textnormal{\predicate{cup-4}}$.


\subsection{Integration of the Inference System}
\label{section:integration}
In order to prevent the curse of dimensionality from propagating from a probabilistic inference system into the perceptual anchoring system, we opt for only loosely integrating the \textit{inference system} with the \textit{anchoring system}. This linkage has as a consequence that we need to maintain two distinct databases for representing our belief of the world.

The above-described anchoring system database plays the role of maintaining a \textit{permanent world model} (PWM), remembering all objects that have appeared over time ($t-k_x$). By contrast, the database of the inference system, implemented in DDC, operates on a \textit{temporary world model} (TWM).
The latter does, however, not only retain which objects are present in the scene but also how the single objects relate to each other. A \predicate{cup} might, for example, be remembered as standing at a certain point in $3{\text-}D$ space but also by the fact that it stands left to some other \predicate{cup}. This representation of the real world is obviously a lot more expensive but adds valuable information to the scene description when carrying out high-level object tracking. The relational nature of the TWM enables us to reason about the world and additionally to track objects on a high level.

By working with two distinct databases we take advantage of the databases for specialized tasks, e.g., reasoning with a relational database. However, it also means that we need to maintain two distinct databases, and more importantly, the databases have to reflect the same world.




Therefore, in order to retain the integrated framework in a coherent state, we need to place the loosely coupled inference system and anchoring system in a tight feedback loop. This feedback loop is represented by the incoming and outgoing arrows in panel~\ftextnumero~3, Figure \ref{fig:system_overview}. It hence consists of two distinct steps:

\begin{enumerate}
	\setlength\itemsep{0.3em} 

	\item Sending anchor information from the \textit{anchoring system} to the \textit{inference system} and initiating or updating the belief of the world in the \textit{inference system}.

	\item Sending back the updated belief of the world in the \textit{inference system} and update the belief of the world in the \textit{anchoring system}, accordingly.

\end{enumerate}

The initial belief in the TWM is initiated by the belief of the PWM of the scene. The anchor information, originating from the PWM, is treated as observations in the TWM. For each initial observation, DDC clauses are added to the TWM database, which constitutes the temporary internal representation of the world. For a \predicate{cup} in the scene, for example, a rule is added that describes the initial belief of its position and velocity:
\begin{flalign}
& \mathtt{pos}(\mathtt{cup})_{0} \sim \mathtt{gaussian}(\vec{\mathtt{R}}, \vec{0}, \vec{\Sigma})\leftarrow \nonumber \\
&\qquad \mathtt{obs}(\mathtt{percept\_pos(cup)})_0\sim = \vec{\mathtt{R}}.&
\label{eq:initial_belief}
\end{flalign}
Where $\vec{\mathtt{X}}$ is the observed $3{\text-}D$ position of the geometric center of the percept (the \predicate{cup} in this case). $\vec{\Sigma}$ is the covariance matrix that specifies the Gaussian and the $\vec{0}$ corresponds to the initial velocity which is set to $0$ in each dimension.

For all the following time steps we define an observation model that takes into account uncertainty in the measurement process itself. We adopt the approach of \citep{nitti2014relational} of expressing the measurement model as the product of Gaussian densities around the position of each object. Assuming independently and identically distributed measurements of the objects allows for this factorization. This idealization assumes that observing an object does not depend on the observation of any other object.
\begin{flalign}
&\mathtt{obs}(\mathtt{percept\_pos
(cup)})_{t+1}\sim \mathtt{gaussian}(\vec{\mathtt{R}}, \Sigma_{obs})\leftarrow \nonumber \\
& \qquad \mathtt{pos}(\mathtt{cup})_t\sim = (\vec{\mathtt{R}}, \mathtt{\_}).&
\label{eq:belief_update}
\end{flalign}
In the case that all objects in the scene are observed, i.e., none of the objects is occluded by another one, the TWM and PWM are now in a state of cognitive consonance, as we used the anchor information as observations for the inference system. If, however, objects get occluded due to manipulations of the world, the occluded objects do not produce any perceptual data anymore. Hence, the perceptual anchoring system can no longer update its belief of the world, and no updated belief is sent to the inference system. In this case, the inference system needs to reason about what might have happened to the object that is not perceived anymore by the anchoring system. Considering the world at time step $t-1$, and the observations of the world at time step $t$, we can speculate about the world in time step $t$. We infer the state of an occluded object through its relations with perceived objects in the world. This inferred updated belief of the world is then sent back to the anchoring system where the state of occluded objects is also updated.

This approach allows us to propose a modified high-level anchoring \textit{track} functionality (cf. \citep{persson.et.al-2017}), such that:

\begin{itemize}
   \setlength\itemsep{0.3em}

	\item \textit{Track} -- extends the definition of an anchor $\alpha^x$ from time $t-1$ to time $t$. This functionality is directly responding to the state of the probabilistic object tracker, which assures that the percepts pointed to by the anchor are the adequate perceptual representation of the object, even though the object is currently not perceived.
  

\end{itemize}

With the introduction of the anchoring \textit{track} functionality, we also need to ensure cognitive consonance at the anchoring side by updating the PWM based on the updated belief of the world, established by the inference system. More specifically, the $3{\text -}D$ \textit{position} attribute $\phi^{pos}_x$ of an anchor $\alpha^x_t$ is updated according to the inferred position of the corresponding object maintained in the TWM of the inference system. This exchange of information between both systems, as described in this section, is further facilitated by sharing the unique identifier $x$ of an anchored object. Hence, we are able to differentiate between specific instances of objects and we can express the rules, given by Eq.~\ref{eq:initial_belief} and \ref{eq:belief_update}, for an object instance that is identified by the unique symbol, e.g., \predicate{cup-1} or \predicate{cup-4}.

The details on how the probabilistic inference is carried out are given in \cite{nitti2013particle, Nitti:2016:PLP:2949339.2949375}. Our contribution lies in coupling a probabilistic inference system with an anchoring system, which enables the conjoined system to probabilistically reason on low-level sensor data. This is for example not the case in \cite{nitti2014relational}, where they used AR-tags to observe objects in the world.



\section{Evaluation and Results}
\label{section:results}

Evaluating a real-world operating anchoring framework, with several interacting components as described in Section~\ref{section:method}, is undoubtedly a challenging task. 
Noisy sensor readings and erroneous attribute measurements are inevitably present and will propagate through the components of the processing pipeline.
The evaluation presented in this section is, therefore, limited to: \textit{1)} the performance of the suggested anchoring matching approach (introduced in Section~\ref{section:learning_anchoring}), and \textit{2)} the integrated combined anchoring and reasoning system (presented in Section~\ref{section:tracking}).

\subsection{Learning the Anchoring Matching Function}
\label{section:learning_anchoring}

The evaluation presented in this section has a two-folded purpose: \textit{1)} collect annotated ground truth data about objects in dynamic scenarios, and \textit{2)} learning to determine which of the the two anchoring functionalities \emph{acquire} or \emph{re-acquire} (cf. Section~\ref{section:anchoring_system} and Figure~\ref{fig:system_overview} -- \ftextnumero~2), to initiate based on the matching distances given of the initial anchoring matching function (given by \cref{eq:caffe_diff,eq:jaccard_index,eq:color_corr,eq:position_diff,eq:time_diff}, as described in Section~\ref{section:matching_function}).

\subsubsection{Data Collection}
\label{section:data_collection}

\begin{figure*}[th!]
	\begin{center}
		\includegraphics[width=0.92\textwidth]{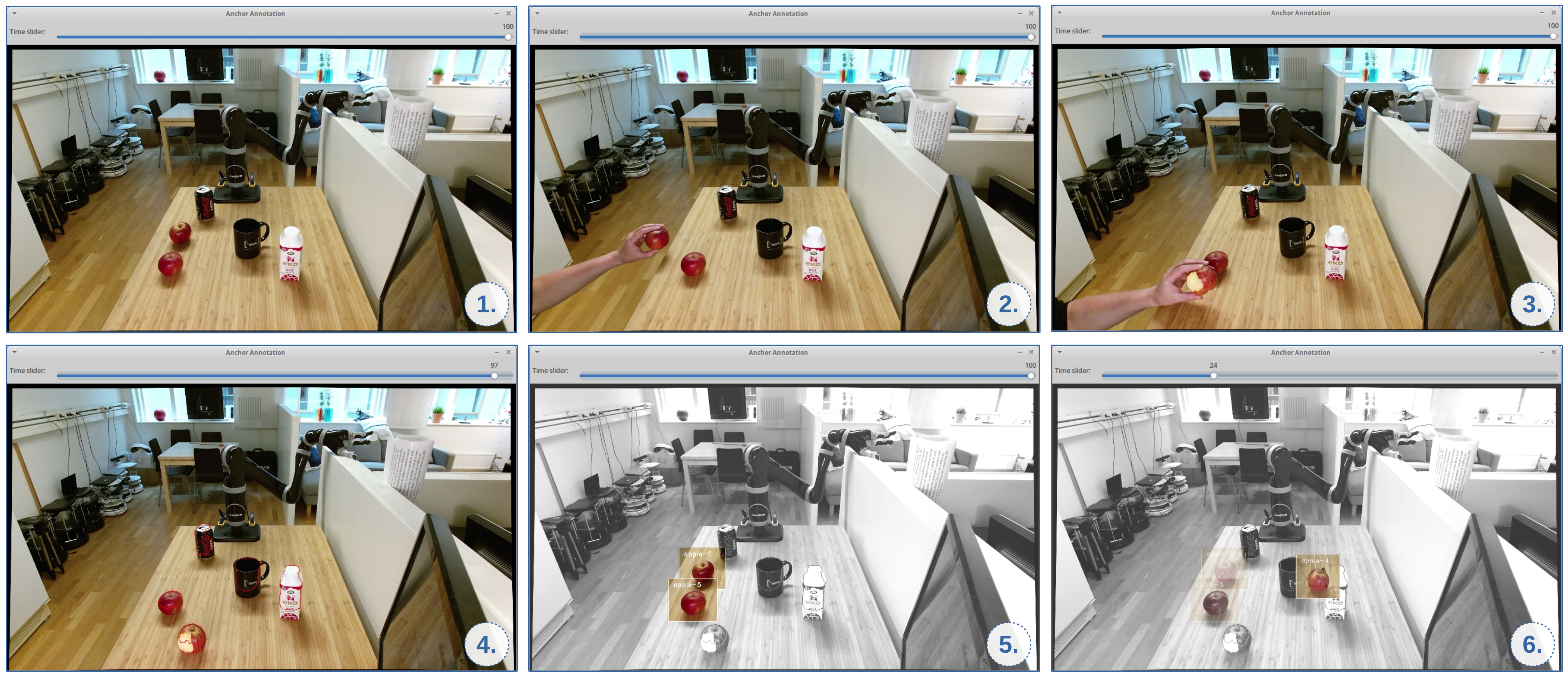}
	\end{center}
	\caption{A depiction of our \textit{human-annotation interface} that was used in order to collect ground truth data of anchored objects. In conjunction with changes in the scene, as illustrated by \ftextnumero~1-3, the human user has the possibility to \textit{freeze} the execution of the framework and providing feedback about what he/she would consider as the appropriate anchoring action for a candidate objects. Once the execution is frozen, the human user can select segmented candidate objects, e.g., the moved \scriptpred{apple} as illustrated in \ftextnumero~4, after which the framework is responding by displaying an updated representation of a number of already anchored objects, shown in \ftextnumero~5, which best (attribute-wise) corresponds to the selected object. The human user can then provide positive feedback about a matching anchored object (by selecting the representation of the matching anchored object), or negative feedback (simply by clicking anywhere else on the screen). Also, to covering the time aspect, and to suggest possible matching anchored objects that have not been perceived recently, we have further added a time slider, illustrated in the top part of \ftextnumero~6. Through this time slider can the user adjust the time factor $k$ for the purpose of selecting the best matching candidate object that was last observed at a time $t-k$. }
    \label{fig:data_collection}
\end{figure*}

A benefit of using perceptual anchoring is that the percepts pointed to by the anchor are the most recent and adequate perceptual representation of an object. For the evaluation presented in this paper, we have exploited the updated and maintained representations, found in anchors, in order to collect human annotated ground truth data. This data collection was conducted through a \textit{human-annotation interface} that was queued with segmented perceptual sensor data given by the perceptual pre-processing pipeline, presented in Section~\ref{section:implementation_etails}. By utilizing this interface, all data about unknown candidate objects, together with the perceptual data of possible matching anchored objects, could be presented and visualized for the human user. Hence, the human was able to provide feedback about the action that the human counterpart would consider as the appropriate anchoring action for each presented candidate object (i.e., \textit{acquire} a new anchor for a queued object, or \textit{re-acquire} an existing anchor). The procedure for collecting our ground truth data is described and exemplified in Figure~\ref{fig:data_collection}.


Behind the scene of proposed human-annotation interface, exemplified in Figure~\ref{fig:data_collection}, the data that in reality was collected and stored was matching similarity scores, provided by \cref{eq:caffe_diff,eq:jaccard_index,eq:color_corr,eq:position_diff,eq:time_diff}. Together with each set of matching similarity scores (as result of comparing the attributes of an unknown candidate object against the attributes of an existing anchored object), was further an annotated label of $1$ stored if the user considered an existing object as a matching object, or $0$ otherwise. Worth noting is that the collected data purely represent that the human user was considering as the most appropriate action for each presented scene. Hence, we were also able to gather samples of objects in, for example, ambiguous situations where an identical (but physically different) instance of an object was introduced while the similar counterpart was still observed. Furthermore, given such ambiguous situations with a number of possible matching anchored objects that could match a selected candidate object, as depicted in Figure~\ref{fig:data_collection} -- \ftextnumero~5-6, we assumed that there could only exist \textit{one} true match (labeled $1$), while the reaming candidates were non-matching candidates (labeled $0$). As a result, we were able to collect several samples for each human action. 


\subsubsection{Experimental Evaluation}
\label{section:experimental_evaluation}

With the use of the human-annotation interface, as described in the previous section, we were able to collect a dataset of a total of $5400$ samples~\footnote{The collected data set is available under http://reground.cs.kuleuven.be and  the \textit{human-annotation interface} is available under bitbucket.org/reground/\\anchoring.}. A dataset that we, subsequently, have used for this particular evaluation in order to train the anchoring system to initiate proper anchoring functionality for different situations. During the data collection, several typical problematic anchoring scenarios, e.g., scenarios there new ambiguous objects are introduced in the scene, scenarios with partly occluded objects, scenarios where existing objects were disappearing and reappearing in the scene, etc., were executed in order to cover a broad range of different situations. Moreover, the data collection was conducted on several occasions for the purpose of capturing changes in the environmental conditions, e.g., changes in light conditions. 

Given the collected data, which was comprised of sets of similarity scores together with corresponding labels (with a label of $1$ for a matching set of similarity scores, or a label of $0$ for a non-matching set), our approach for learning how to correctly anchoring objects, and thereby learn to invoke correct anchoring functionality (\textit{acquire} or \textit{re-acquire}), was through the evaluation of different classification algorithms. More specifically, for this evaluation we have tested and trained following classification algorithms: 

\begin{itemize}
    \setlength\itemsep{0.3em}
	
    \item Support Vector Machine (SVM)\cite{burges-1998}, with $\nu{\text -}$Support Vectors (trained with $\nu=0.1$), and with a \textit{ Histogram intersection} kernel function
    
    \item Multi Layer Perceptron (MLP), with \textit{back-propagation} training, two hidden layers and a layer configuration, according to: $x-10-15-2$
  	
    \item k-Nearest Neighbor (k-NN), trained and tested with $k=3$  
    
    \item Normal Bayes Classifier (Bayes)\cite{fukunaga-2013}
    

\end{itemize}

\begin{figure}[th!]
	\begin{center}
		\includegraphics[width=0.48\textwidth]{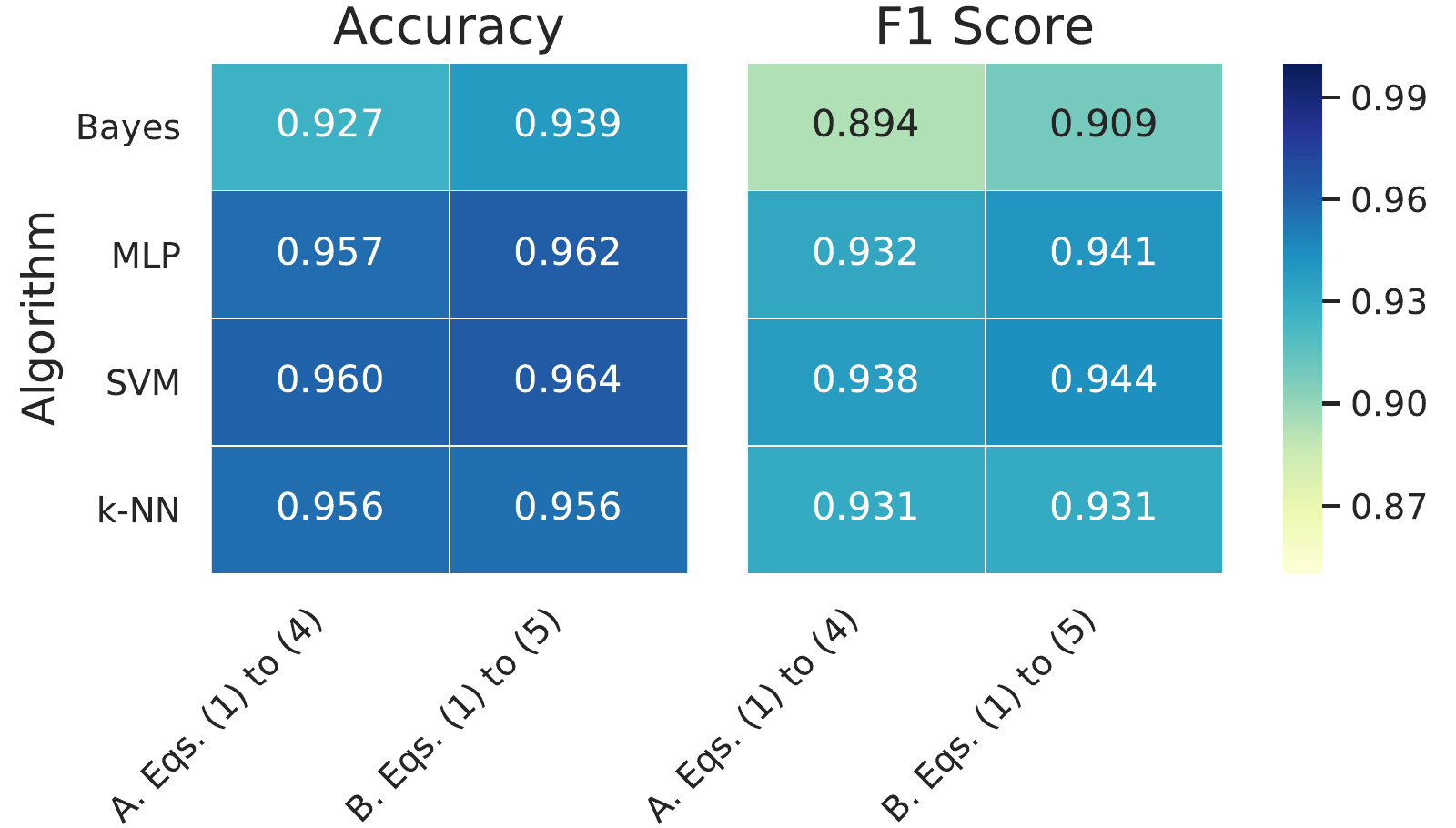}
	\end{center}
	\caption{Resulting average \textit{classification accuracy} together with \textit{F1 score} for each used model for our approach to learn the anchoring functionalities. }
    \label{fig:anchoring_learning}
\end{figure}

Collected dataset was randomly divided $70/30$ into training/test samples, giving us a total of $3780$ \textit{training} samples and $1620$ \textit{testing} samples. Resulting average classification accuracy of each trained approach is listed in Figure~\ref{fig:anchoring_learning}.


Given the result, presented in Figure~\ref{fig:anchoring_learning}, it is seen that the best \textit{average classification accuracy} of $96.4\%$ was achieved by the use of the SVM classifier. The highest average \textit{F1 score} (for a \textit{true match}) of $94.4\%$ was, likewise, achieved with the same SVM classifier. By the results seen in Figure~\ref{fig:anchoring_learning}, it should, however, also be noted that the differences in accuracy between the MLP classifier and the SVM classifier are close to insignificant (only $0.2\%$). Nevertheless, the best trained resulting SVM model was formally integrated as a part of the initial \textit{matching function} of the anchoring system such that the predicted result of the SVM model was used to determine if an unknown candidate object was matching an existing anchor (i.e., if the object should be \textit{re-acquired} as an existing matching anchor), or if no current anchors were matching the candidate object (i.e., if a new anchor should be \textit{acquired} for the object). Integrated classification approach was, subsequently, used for the remaining experiments presented in this paper. 

By comparing the results between omitting (\textit{column A}) or considering (\textit{column B}) the time difference as an additional feature, mapped to a time distance according to \cref{eq:time_diff}, it is also evident that the time $t$, in fact, is a relevant factor for the concept of anchoring. The intuition behind including this additional time feature in the evaluations presented in this section was to capture time-dependent changes in the environment while learning how to anchor objects, e.g., the position of an object can only change with a limited velocity between sequential frames. Through examining the results for the best resulting SVM models, it is seen that our intuition was correct and that we achieved $0.4\%$ better classification accuracy and $0.6\%$ better F1 score, as a result of including the time difference as a feature. Worth noted, it can further be seen by the results in Figure~\ref{fig:anchoring_learning}, that the k-NN classifier was the only classifier that did not benefit from increasing the dimensionality of the input data by including the time difference as a feature.

Finally, it should be noted that the integrated classification approach can, in some cases, returning several matching candidate anchors (i.e., a candidate object can be \textit{re-acquired} as more than one existing matching anchor). It is, therefore, important to globally consider all possible candidates for all observed objects in each frame in order to determine the best matching candidate anchor for each observed object. For the work presented in this paper, we used an SVM classifier with continuous output values such that the globally best matching candidate anchor was determined in a \textit{winner takes all} manner. 


\subsection{Tracking of Occluded Objects}
\label{section:tracking}

Despite the accuracy of the anchoring system, presented in previously Section~\ref{section:learning_anchoring}, there are scenarios where the pure anchoring system fails to correctly \textit{acquire} (or \textit{re-acquire}) an object, e.g., when an object is occluded and moved by another object. For a changing and adaptable system to handle the world modeling of such scenarios, the system must further incorporate \textit{model-based object tracking} in order to maintain  objects that are not perceived by the input sensors.
In this section, we will exemplify how our approach of integrating DDC into the anchoring system (cf. Section~\ref{section:integration} and Figure~\ref{fig:system_overview} -- \ftextnumero~2\&3) can handle such scenarios with occluded objects, and as a subsequent result further improve the anchoring accuracy.  

\subsubsection{Proof of Concept}
\label{section:proof_of_concept}

\begin{figure*}[ht!]
	\begin{center}
		\includegraphics[width=0.92\textwidth]{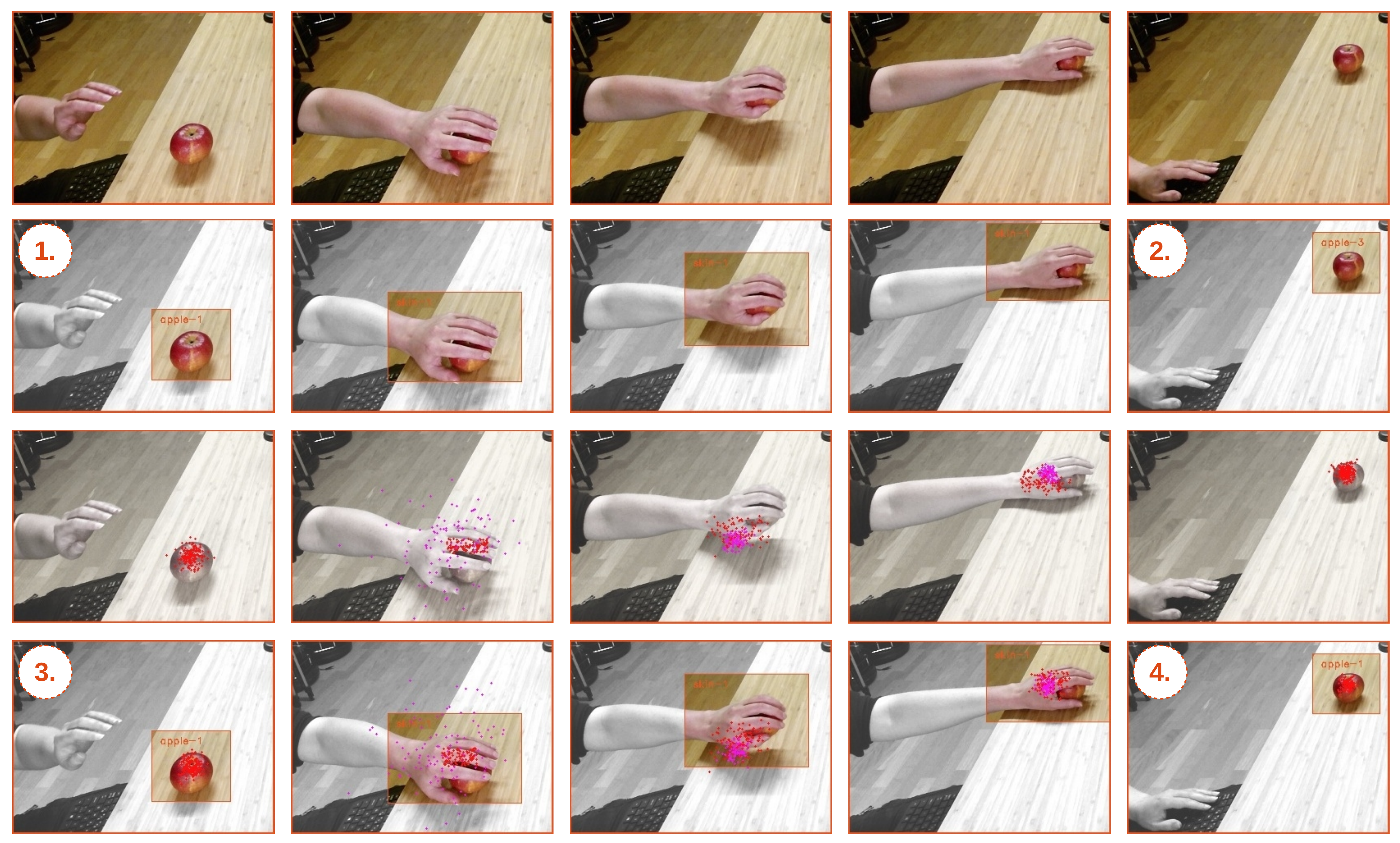}
	\end{center}
	\caption{A depiction of how suggested system benefits of combined object anchoring and probabilistic object tracking. \textit{Rows in order from the top:} \textit{1\textsuperscript{st})} representing screen-shots of a scenario where a human hand is occluding an apple while the apple is moved, \textit{2\textsuperscript{nd})} corresponding resulting anchored objects while \textit{only} using the \textit{anchoring system} (note that the original \scriptpred{apple-1} object is lost while it is occluded and moved by the \scriptpred{skin-1} object, and a new \scriptpred{apple-3} object is, therefore, \textit{acquired} in the end of the scenario), \textit{3\textsuperscript{rd})} plotted particles given by the \textit{inference system} during execution of suggested integrated approach, and \textit{4\textsuperscript{th})} corresponding resulting anchored objects of the \textit{anchoring system} supported by the feed back of the \textit{inference system} (note that in this case is the position of \scriptpred{apple-1} object tracked while it is occluded and moved by the \scriptpred{skin-1} object, and the \scriptpred{apple-1} object is, accordingly, \textit{re-acquired} in the end of the scenario).
    }
    \label{fig:proof_of_concept}
\end{figure*}

To demonstrate how the anchoring system benefits from the feedback of the inference system (and consequently how the inference system benefits from the same integration), we plot the particles in form of point positions (representing the belief of the world in the inference system, as described in Section~\ref{section:integration}), concurrently with the output of the anchoring system, as exemplified in Figure~\ref{fig:proof_of_concept}. The mean position in $3{\text -}D$ space of the particles for each object that was not directly perceived at time $t$, e.g., objects occluded by another object, was subsequently fed back to the reinstated \textit{track} functionality of the \textit{anchoring system} such that the position of an anchor (even though  the anchored object was not observed), was updated to the most probable position according to the \textit{inference system}. 

Comparing the resulting anchored objects, seen in Figure~\ref{fig:proof_of_concept}, it is evident that there is a significant difference in resulting anchors. In the case where only the \textit{anchoring system} was used (Figure~\ref{fig:proof_of_concept} -- \textit{2\textsuperscript{nd} row from top}), it can be seen that the initial \predicate{apple-1} object (seen in Figure~\ref{fig:proof_of_concept} -- \ftextnumero~1) is lost while the object is occluded and moved by the \predicate{skin-1} object. Consequently, when the apple object reappears in the scene, the anchoring system cannot determine if the object is a new \predicate{apple} or the previously anchored \predicate{apple-1}, and as a result \textit{acquire} an new anchor \predicate{apple-3} (seen in Figure~\ref{fig:proof_of_concept} -- \ftextnumero~2). However, in the case where both the \textit{anchoring system} and the \textit{inference system} are used (Figure~\ref{fig:proof_of_concept} -- \textit{bottom row}), and where the position of the \textit{tracked} \predicate{apple-1} object (seen in Figure~\ref{fig:proof_of_concept} -- \ftextnumero~3) is fed back to anchoring system while the object is moved, it can seen that the apple object instead is correctly \textit{re-acquired} as \predicate{apple-1} once the object reappears in the scene (seen in Figure~\ref{fig:proof_of_concept} -- \ftextnumero~4). Note that rather than using a dedicated classifier for recognizing different human body parts, we have, instead, fine-tuned our GoogLeNet model to recognize human \scriptpred{skin} objects as one of the object categories.

\subsubsection{Exemplifying Scenarios}
\label{section:scenarios}

Given the exemplified proof-of-concept (presented in the previous Section~\ref{section:proof_of_concept}), we will in this section further demonstrate a number of scenarios where our suggested integrated system excels (compared to an anchoring approach that exclusively is based on perceptual observations of objects):   

\begin{enumerate}
    \setlength\itemsep{0.3em}

    \item \textit{Simple occlusion} -- we start with two objects (among other objects) that are both visible. We then hide the smaller one of the objects behind the bigger one. The occluded object does not produce any sensor data. We can, however, reason about it. Then the smaller object reappears in the scene, we should be able to associate the reappearing object with the one from before (same anchor with high probability).
    
    
    \item \textit{Moving an occluded object} -- we now want to \textit{track} an object for which no sensor data is available. We start again with two objects that are both visible. We then hide the smaller object underneath the bigger object, move the bigger object and, subsequently, reveal the smaller object. We should be able to associate the reappearing object with the one from before.

    
    \item \textit{Moving occluded objects with unexpected revealing} -- similar scenario as before only this time we start out with also having an unknown object hidden which the observer initially does not know about. We now hide again the (visible) smaller object underneath the bigger object, move the bigger object, but this time reveal the initially unknown hidden object. The system should recognize this as a new object. Then the other (initially visible) smaller object is revealed, the system should recognize this object as the previously anchored object.
    
    \item \textit{Shell game} -- we start with three identical containers and a smaller object. We then hide the smaller object underneath one of the three containers and start shuffling the containers around. We should now be able to ask the system under which of the three containers the hidden objects is located.
    
\end{enumerate}

\begin{figure*}[ht!]
	\begin{center}
		\includegraphics[width=0.92\textwidth]{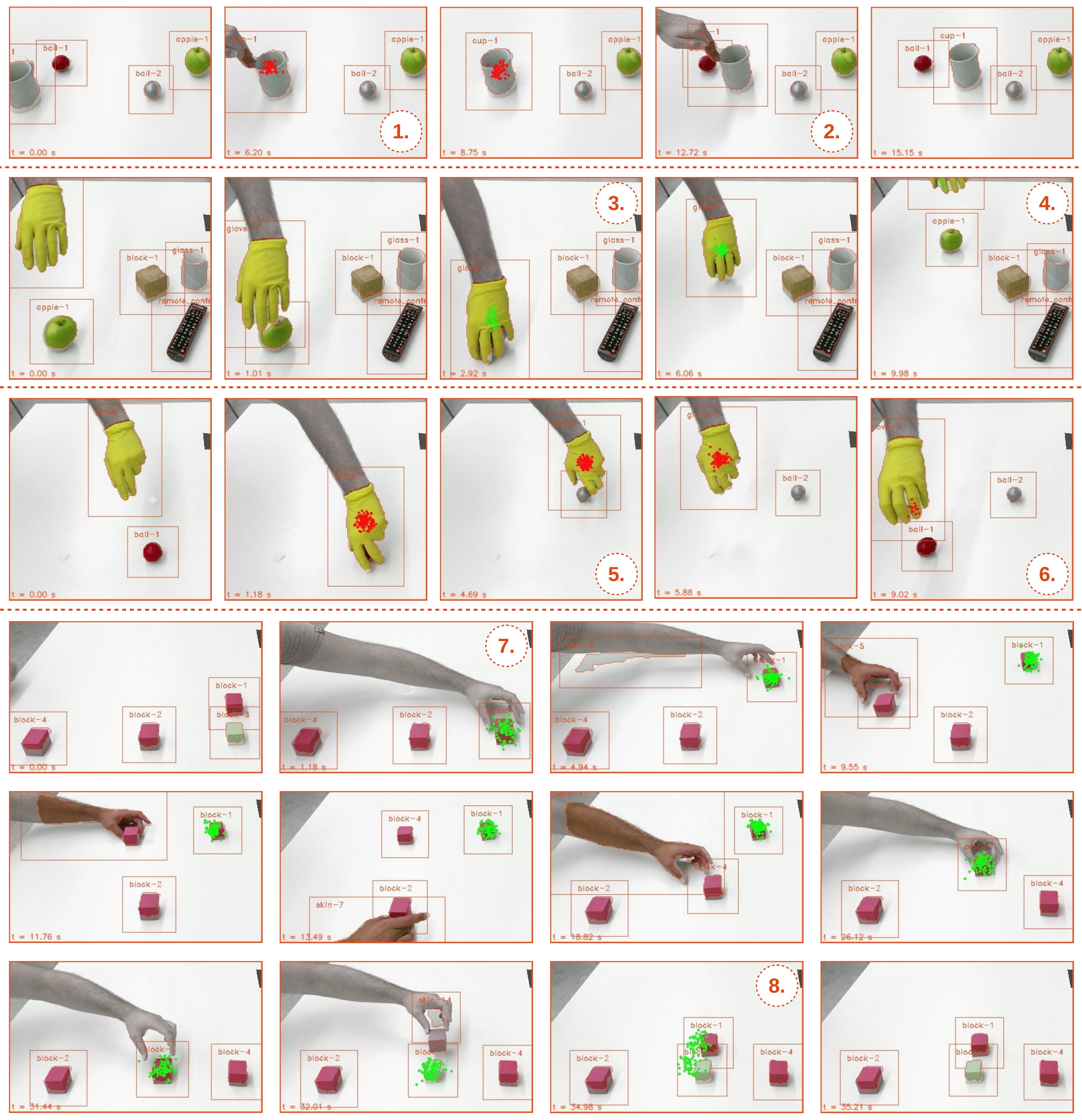}
	\end{center}
	\caption{ Examples of screen-shots captured during the execution of stated scenarios. Visual perceived anchored objects are symbolized with the unique anchor id (e.g., \scriptpred{ball-2}), while occluded hidden objects are depicted by plotted particles that represent possible positions of the occluded object in the inference system. \textit{Rows in order from the top:} \textit{1\textsuperscript{st})} example of \textit{simple occlusion} where a \scriptpred{ball} is hidden behind a \scriptpred{cup}, \textit{2\textsuperscript{nd})} depicts the \textit{movement} of an \textit{occluded object} where a \scriptpred{glove} (or human hand) is occluding while moving an \scriptpred{apple}, \textit{3\textsuperscript{rd})} similar example of \textit{moving an occluded object} where a \scriptpred{glove} is occluding while moving a \scriptpred{ball} (\scriptpred{ball-1}), but in this case is also another \scriptpred{ball} object (\scriptpred{ball-2}) introduced during the execution of the scenario, \textit{4-6\textsuperscript{th})} illustrate a \textit{shell game} scenario where a smaller object (\scriptpred{block-3}) is hidden under one of three identical containers (\scriptpred{block-2}), and where the containers, subsequently, are shuffled around.  
    }
    \label{fig:scenarios}
\end{figure*}

In Figure~\ref{fig:scenarios}, we exemplify our results of stated scenarios with a number of screen-shots during the execution of each scenario.\footnote{Full videos are available under http://reground.cs.kuleuven.be/}. The scenarios were performed in near real-time on a laptop Intel(R) i7 CPU 2.60GHz with 16 GB memory and an NVIDIA Quadro M1000M. This constitutes a promising feature of our approach as we do not need access to high-performance machines to deploy our system.

In the first example with \textit{simple occlusion} (Figure~\ref{fig:scenarios} -- \textit{1\textsuperscript{st} row from top}), it can be seen that as soon as the  \predicate{cup} occludes the smaller \predicate{ball} object, the object is immediately tracked and maintained by the probabilistic reasoner (seen in Figure~\ref{fig:scenarios} -- \ftextnumero~1). Opposite, once the \predicate{ball} objects reappear in the scene, it is no longer any need to probabilistically track the object, and the object is, once again, maintained through anchoring (seen in Figure~\ref{fig:scenarios} -- \ftextnumero~2).

Through the second example (Figure~\ref{fig:scenarios} -- \textit{2\textsuperscript{nd} row from top}), it is illustrated how the combined system handles \textit{movements during occlusions}. In this example, a \predicate{glove} object (a human hand) is occluding while moving an \predicate{apple} object. As soon as the \predicate{apple} object is occluded by the \predicate{glove}, the \predicate{apple} object is tracked and maintained though probabilistic reasoning (seen in Figure~\ref{fig:scenarios} -- \ftextnumero~3). The tracked position of the occluded object is continuously fed back to the anchoring system (through the newly instated anchoring \textit{track} functionality), and the \predicate{apple} object is, consequently, \textit{re-acquired} as the same \predicate{apple-1} once the object reappears in the scene (seen in Figure~\ref{fig:scenarios} -- \ftextnumero~4). 

In the third example (Figure~\ref{fig:scenarios} -- \textit{3\textsuperscript{rd} row from top}), we demonstrate how our combined systems truly works in symbiosis. In this case, a similar scenario of \textit{moving an occluded object} is exemplified where a \predicate{ball} (\predicate{ball-1}) is occluded while moved by a \predicate{glove}. However, another unknown \predicate{ball} is initially also hidden underneath the \predicate{glove} (in the human hand). This hidden \predicate{ball} is later introduced in the scene during the execution of the scenario (seen in Figure~\ref{fig:scenarios} -- \ftextnumero~5). Nevertheless, since the second \predicate{ball} is different in appearance (compared to \predicate{ball-1}), this newly introduced object is correctly \textit{acquired} as an new \predicate{ball-2} object, while the first \predicate{ball-1} object correctly remains tracked, and is subsequently \textit{re-acquired} as the same \predicate{ball-1} object once reappearing in the scene (seen in Figure~\ref{fig:scenarios} -- \ftextnumero~6).

Finally, in the fourth example (Figure~\ref{fig:scenarios} -- \textit{4\textsuperscript{th}-6\textsuperscript{th} row from top}), we reconnect with our initial motivation statement (outlined in Section~\ref{section:introduction}), through presented screen-shots captured during execution of the \textit{shell game} scenario.
In this example, a smaller \predicate{block} object (\predicate{block-3}) is hidden underneath one of three identical larger \predicate{block} objects (\predicate{block-1}), seen in Figure~\ref{fig:scenarios} -- \ftextnumero~7. All the larger \predicate{block} objects are, subsequently, moved around and shuffled. Nevertheless, during all movements is the hidden \predicate{block-3} object tracked though the relation with the occluding counterpart (\predicate{block-1}), i.e., the inference system is repeatedly speculating about the position of the hidden \predicate{block-3}, and the tracked position of is continuously fed back to the anchoring system. Consequently, once the hidden object is revealed and reappear in the scene (seen in Figure~\ref{fig:scenarios} -- \ftextnumero~8), the object is correctly \textit{re-acquired} as the same \predicate{block-3} object.


\section{Conclusions and Future Work}
\label{section:conclusions}

In this paper, we have presented how we are able to improve the overall anchoring process by introducing a post-anchoring high-level probabilistic reasoning procedure with the purpose of predicting the state of objects that are not directly perceived through the perceptual sensor data, e.g., in case of object occlusions. To retain the integrated framework in a coherent cognitive state, we have suggested a loosely coupled integration between proposed inference system and anchoring system, while a tight feedback loop is preserved in order to maintain consented tracked positions of objects. We have presented the proof-of-concept of how this integrated framework is used to model and manage a consistent semantic world model of perceived objects in dynamic scenarios. We have further introduced a novel anchoring matching approach based on classification of humanly annotated ground truth data of real-world objects for determining whether a perceived object has previously been observed (or not), and, subsequently, invoke correct anchoring functionality (\textit{acquire} or \textit{re-acquire}), in order to correctly anchor perceived objects. Through the presented results, we have shown that our learned anchoring matching approach is able to accurately anchoring objects and maintaining consistent representations of objects with an accuracy of $96.4\%$.


In the scope of the ReGROUND project (cf. Section~\ref{section:introduction}), a possible future direction is to exploit how anchored objects and their spatial relationship, tracked over time, facilitate the learning of both the function of objects, as well as object affordances -- similar to previously presented works on learning objects affordances from visual data \cite{kjellstrom.et.al-2011,koppula.et.al-2013,koppula&saxena-2014}. However, previously presented works have commonly assumed an approach based on the tracking of human hand actions, e.g., exploiting the spatial-temporal relationships between objects and human hand actions to learn the function of objects \cite{kjellstrom.et.al-2011}. For the approach presented in this paper, we only consider the tracking of object instances (where a human \predicate{hand} object might constitute one such object instance), and where a relational particle filter approach is utilized for high-level tracking. Hence, it is also possible to further infer both the function of objects and object affordances by employing probabilistic logic programming.

The integration of high-level object tracking into the anchoring framework is also a subject for further investigation. As of now, only uni-modal probability distribution can be handled by the anchoring system. This means that the rich and intricate probability distributions that can be expressed in DDC, for example multi-modal distribution of the position of unobserved objects, can not be passed on to the anchoring system and be handled in a probabilistic fashion. Allowing this would render our approach truly probabilistic and equally allow us to keep track of multiple hypotheses.  

\section*{Acknowledgments}


This work has been supported by the ReGROUND project (http://reground.cs.kuleuven.be), which is a  CHIST-ERA project funded by the EU H2020 framework program, the Research Foundation - Flanders and the Swedish Research Council (Vetenskapsr{\aa}det). The work is also supported by the Swedish Research Council (Vetenskapsr{\aa}det) under the grant number: 2016-05321. The authors would also like to thank Prof. Alessandro Saffiotti for his valuable comments and discussions during the work.


\bibliographystyle{IEEEtranN}
\bibliography{references.bib}


\end{document}